\documentclass[preprint,11pt]{elsarticle}
\usepackage[utf8]{inputenc}
\usepackage{amssymb}
\usepackage{multirow}
\usepackage{color}
\usepackage{amsmath}
\usepackage{amsthm}
\usepackage{rotating}
\usepackage{graphicx}
\usepackage{nicefrac}
\usepackage[hidelinks]{hyperref}
\usepackage{algpseudocode}
\usepackage{lscape}
\usepackage{booktabs}
\usepackage[ruled,vlined,linesnumbered]{algorithm2e}

\title{The two-echelon routing problem with truck and drones}
\begin{document}

\begin{center}
\begin{LARGE}
The two-echelon routing problem with truck and drones
\end{LARGE}

\vspace*{01cm}
\textbf{Minh Ho\`ang H\`a*,  Lam Vu, Duy Manh Vu} \\
ORLab, Faculty of Computer Science, Phenikaa University, Hanoi, VietNam\\

%144 Xuan Thuy, Cau Giay, Hanoi, Vietnam \\

\vspace*{0.2cm}

\end{center}

\noindent
\textbf{Abstract.} In this paper, we study novel variants of the well-known two-echelon vehicle routing problem in which a truck works on the first echelon to transport parcels and a fleet of drones to intermediate depots while in the second echelon, the drones are used to deliver parcels from intermediate depots to customers. The objective is to minimize the completion time instead of the transportation cost as in classical 2-echelon vehicle routing problems. Depending on the context, a drone can be launched from the truck at an intermediate depot once (single trip drone) or several times (multiple trip drone). Mixed Integer Linear Programming (MILP) models are first proposed to formulate mathematically the problems and solve to optimality small-size instances. To handle larger instances, a metaheuristic based on the idea of Greedy Randomized Adaptive Search Procedure (GRASP) is introduced. Experimental results obtained on instances of different contexts are reported and analyzed.
\vspace*{0.2cm}


\section{Introduction}
\label{intro}
Nowadays, as technology for aerial mobility progresses, delivery schemes are naturally destined to make use of unmanned aerial vehicles (UAVs or drones), and multiple companies such as Amazon \cite{amazon}, UPS \cite{ups}, DHL \cite{dhl}, 7-Eleven \cite{7Eleven} and Alibaba \cite{alibaba} have already prepared for this transition in practice. There are four advantages of using drones for delivery: (1) it can be operated without a human driver, (2) it flies over traditional road networks and as a result can visit truck-unreachable customers, (3) it is faster than trucks, and (4) it has much lower transportation costs per kilometer. Yet, drones also pose specific challenges, due to their limited autonomy, smaller capacity, and possible need of visiting stations for charging or swapping batteries. To offset the drawbacks of both vehicles, they are coupled together to make delivery as proposed in multiple recently published works \cite{murray2015flying, ha2018min}. In this paper, we consider situations in which the customers cannot be reached by heavy truck and must be serviced by drones. This raises new routing variants that use the combination of truck and drones for delivery. We call the problems as Two-echelon Routing Problems with Truck and Drones (2ER-TD) and define them as prototypical problems. The main difference between the 2ER-TD and the classical Two-Echelon Vehicle Routing Problem (2EVRP) is that drones are used in the second echelon instead of trucks. In addition, because the drones are transported by the truck to intermediate depots, the truck after launching the drones must stay at an intermediate depot to wait for the drones coming back.

The 2ER-TD is a generalization of the traveling salesman problem (TSP) and can be defined formally as follows. Let $G = (V \cup W, E_1 \cup E2)$ be an undirected graph, where $V \cup W$ is the node set and $E_1 \cup E_2$ is the edge set. $V = \{v_0,..., v_{n-1}\}$ is the set of $n$ nodes that can be visited by truck, and $W = \{w_1,w_2,..., w_l\}$ is the set of nodes that must be delivered by drones. The nodes in $V$ and $W$ are also called truck nodes and drone nodes (or customers), respectively. Node $v_0$ is the depot; a truck and a set $U$ of $u$ identical drones are located there. Each drone has a flying range $r$ measured by the time drone can autonomously fly. The truck does not service customers directly but plays the role of a mobile depot to help the drones reaching customers to deliver. Because the drones travel between nodes in $V$ with the truck, after the drones are launched from the truck at a node $v_i \in V$, the truck stays at $i$ to wait for the drones coming back. A travel time $t_{ij}$ is associated with each edge of $E_1 = \{(v_i, v_j): v_i, v_j \in V, i < j\}$ and a travel time $t'_{ij}$ is associated with each edge of $E_2 = \{(v_i, v_j): v_i \in V, v_j \in W\}$. The truck travels along the edges of $E_1$ while drones perform back and forth trips along the edges of $E_2$ to service customers. Assuming that both the truck and the drones start from the depot at time 0. A drone can service only one customer during a flight. The objective of the 2ER-TD is to minimize the delivery completion time, that is, the time at which the truck and all drones are back to the depot, with the service of all customers being complete.

\begin{figure}[ht]
\centering
\includegraphics[scale = 0.21]{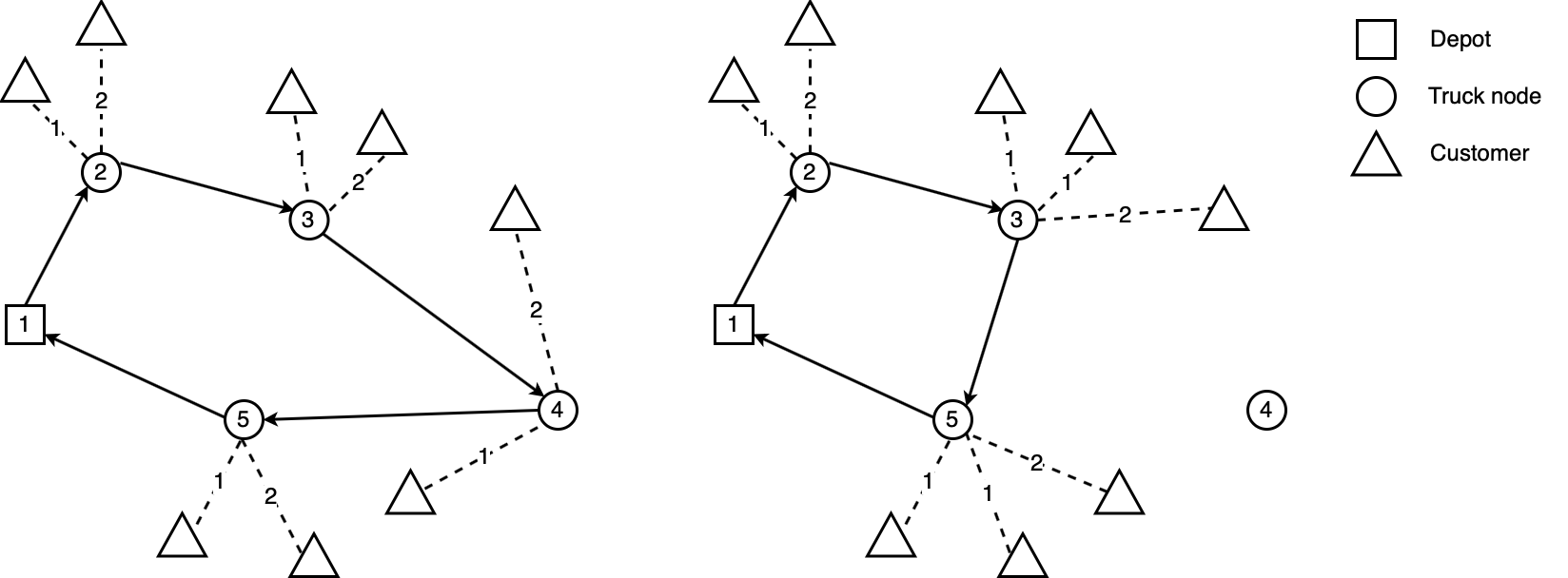}
\caption{An example of solutions for 2ER-TDs (left) and 2ER-TDm (right)}
\label{fig1}
\end{figure}

Two variants can arise in practice: the 2ER-TD with single trip drone (2ER-TDs) and the 2ER-TD with multiple trip drone (2ER-TDm). At any node $v_i \in V$, the former considers a drone performs only one flight to service a customer while in the latter, a drone can perform multiple back and forth trips to service many customers. Obviously, the second variant is more general. However, the first one can arise in several contexts for some reasons. First, when there are relatively enough drones, a drone does not have to perform multiple trips. Second, the scheduler wants to keep routing plans simple and does not have to care about the delivery order for each drone. Third, it could be unnecessary to spend time on loading parcels into the drones while the truck is stationary to wait for the drones coming back. Instead, preparing parcels can be done while the truck travels on the road to save more time. Figure \ref{fig1} illustrates these two situations. The dotted lines show the back-and-forth drone trips while the solid ones belong to the truck trip. The numbers on the dotted lines represent the drone identification. For example, from truck node 3 of the 2ER-TDm solution, drone 1 performs two trips while drone 2 performs one trip. As shown in Figure \ref{fig1}, allowing the drones to perform multiple trips can reduce the number of nodes visited by the truck (the truck node 4 is unnecessary to visit in the 2ER-TDm solution).

The problem has a potential application in humanitarian logistics to solve disaster relief problems. In this application, after a disaster, health care organizations have to supply the affected populations with food, water, and medicine. Due to road's condition, the relief vehicle (e.g. mobile hospital) cannot approach affected areas and must stop at several locations from which drones are launched to deliver products to the isolated populations (the set $W$ in the mathematical description of the problem). The health care organizations have to choose the appropriate stops among $|V|$ potential locations so that all populations can be serviced as soon as possible. It is worth mentioning that, in case the demand of an affected population exceeds the workload of a drone, a fleet of drones is required to fulfill the demand of a customer. This situation can still be covered by our problem models by multiplying a customer into many copies with the same location such that a customer copy can be serviced by a single drone. 

Another application of the problem is to deliver commercial parcels in areas of high density customers. In this context, a number of drones can be launched at an intermediate depot to service many customers. Due to the large number of drones, the scheduler wants to keep the system as simple as possible by not allowing the truck to launch and retrieve the drones at different intermediate depots as proposed in \cite{ha2018min, murray2015flying}. This transportation system can increase the service level when it services the customers faster than the classical TSP-based method. In the experimental section, we will analyse and compare these two transportation modes in terms of the service completion time, i.e. the time coming back at the depot of vehicles.

The contributions of our research are as follows. First, we introduce two new variants of routing problems with drones. Second, we formulate the problems as Mixed Linear Integer Programs (MLIP) from which we can solve small-size instances to optimality. Third, a metaheuristic based on the idea of Greedy Randomized Adaptive Search Procedure (GRASP) with new solution construction and local search operators is proposed. Finally, we introduce first benchmark instances to investigate the hardness of the problems and the performance of the solution methods, as well as to provide other insightful conclusions about the system. In particular, we show that the transportation model based on the 2ER-TD can be applied in not only humanitarian logistics but also commercial delivery to increase the service quality.

The remainder of the paper is structured as follows. Section \ref{review} reviews related works while Section \ref{formu} describes our formulations. Our metaheuristic is presented in Section \ref{meta}. Finally, Section \ref{result} discusses the computational results, and Section \ref{conclu} summarizes our conclusions.

\section{Literature review}
\label{review}
The 2ER-TV is related to three problems well studied in the literature: the two-echelon vehicle routing problems \cite{Perboli2011}, the covering salesman problem, and the routing problems with drones. We now review the main studies on these problems.

\textbf{Two-echelon vehicle routing problems}. There are two main differences between the 2ER-TD and the classical two-echelon vehicle routing problems (2EVRP). First, the transportation in the second echelon of the 2ER-TD is operated by the drones, not by trucks as in the 2EVRP. Compared with the truck, the drone has limited flight range and performs back and forth trips to service only one customer during a flight. Second, the objective function in the 2ER-TD is minimizing the time the truck and drones coming back to the depot. This leads to the fact that, at the intermediate depots, the truck must wait for all the drones returning. While in the 2EVRP, there is no need for the trucks in the first echelon waiting for the trucks in the second echelon coming back because the objective function is minimizing the total transportation cost of the whole system.

Several early studies on 2EVRP focus on exact methods based on mathematical programming solution techniques. The first branch-and-cut algorithm for the 2EVRP is described in the PhD thesis of Gonzalez-Feliu \cite{Feliu2008}. Based on a commodity flow formulation, the algorithm can solve instances with up to 32 customers and 2 satellites. Other branch-and-cut algorithms can be found in \cite{Perboli2010, Perboli2011, Jepsen2013, Perboli2018}. Santos et al. \cite{Santos2015} describe a branch-and-cut-and-price algorithm for the 2EVRP that relies on a reformulation based on the $q$-routes relaxation. Baldacci et al. \cite{Baldacci2013} propose an exact method based on a set partitioning formulation with side constraints. The results obtained on 207 instances with up to 100 customers and 6 satellites show that their exact algorithm outperforms the existing methods from the papers described above. The number of metaheuristics that solve the 2EVRP has also rapidly grown in the last decade. We refer the readers to the surveys by \cite{Cuda2015, Schiffer2019} for further information about this line of research. Last but not the least, the studies on 2EVRP have recently been extended to the contexts where electric vehicles (which typically have a larger autonomy than drones) are used in the second echelon (see, for example, \cite{Breunig2019, Jie2019}). 

\textbf{Covering salesman problems.} The class of covering salesman problems (CSP) is also similar to 2ER-TV in the feature that we also have to determine the location of intermediate depots. However, the basic difference lies in the second echelon where it is not necessary to deliver parcels directly from the intermediate depots to customer locations. Instead, the customers are required to travel to intermediate depots to pick up parcels. This leads to the covering constraints. That is, we need to find the location for intermediate depots such that every customer is covered at least once, i.e. being within some predefined distance from at least one intermediate depot. Moreover, this class of the problem often considers the objective function that minimizes the transportation cost, not the service completion time as in 2ER-TD. The covering salesman problem is first proposed in \cite{current1989}. It is studied in more detail by a branch-and-cut algorithm and a heuristic in \cite{gendreau1997}. The version with multiple vehicles is solved more intensively by a branch-and-cut algorithm and a heuristic in \cite{ha2012}, a branch-and-price algorithm in \cite{Jozefowiez2014}, a variable neighborhood search (VNS) in \cite{Kammoun2017}. The variants in which a customer must be covered at least not only once but several times are considered in \cite{golden2012, pham2017}. Basically, the 2ER-TV is more difficult than CSPs due to additional decision layers required to be handled in the second echelon. Therefore, the solution methods proposed for the CSPs cannot be applied directly to solve the 2ER-TV. 

\textbf{Routing problems with trucks and drones}. The model of combining trucks and drones is first introduced by Murray and Chu (2015) \cite{murray2015flying} as the Flying Sidekick Traveling Salesman Problem (FSTSP). In this proposed model, a drone is attached on top of the delivery truck, simultaneously moves along with the truck, and flies from the truck to make a delivery while the truck continues to service its customers in different locations. Once the drone completes its service for one customer, it needs to fly back to the truck at the current delivery location or along the route to the next delivery location. This problem differs from the 2ER-TD in the synchronization operation between truck and drone. In FSTSP, while waiting for the drone, the truck itself can travel along the way to service other customers. On the other hand, in the 2ER-TD, the truck remains stationary to wait for the drone coming back and cannot service any customer. In addition, the FSTSP makes routing schedules for only one drone while in 2ER-TD, multiple drones are considered.

In \cite{murray2015flying}, a mixed integer linear program is proposed to formally define the FSTSP. A simple and fast heuristic is also designed to solve instances with up to 10 customers. The Traveling Salesman Problem with Drone (TSP-D) in \cite{ha2018min} is defined similarly to the FSTSP except for the objective function of minimizing the travel cost rather than the latest time that both vehicles return to the depot. Two algorithms are designed to solve the TSP-D. The first algorithm is adapted from the idea of the approach proposed by \cite{murray2015flying}, in which an optimal TSP solution is converted to a feasible TSP-D solution by local searches. The second algorithm, a Greedy Randomized Adaptive Search Procedure (GRASP), is based on a new split procedure that optimally splits any TSP tour into a TSP-D solution. After a TSP-D solution has been generated, it is then improved through local search operators. Freitas et al. \cite{Freitas} introduce a hybrid heuristic named HGVNS to solve the FSTSP. The algorithm first generates the initial solution by using a MILP solver to solve the TSP optimally and then applies a heuristic in which some trucks’ customers are removed and reinserted as drone customers. Next, the initial solution is used as the input for a general variable neighbourhood search in which eight neighbourhoods are shuffled and chosen randomly. In a more recent work, Ha et al. \cite{Minh2020} propose a Hybrid Genetic Search (HGS) with dynamic population management and adaptive diversity control to solve the TSP-D problems under both min-cost and min-time objectives. Based on problem-tailored components such as a split algorithm, crossovers, local search operators, and an adaptive penalization mechanism, the HGS is able to dynamically balance the search between feasible/infeasible solutions. The computational results show that the proposed algorithm outperforms two existing methods of \cite{Freitas, ha2018min} in terms of solution quality. Other papers, including Agatz et al. \cite{AgatzBS18} and Poikonen et al. \cite{PoikonenGW19} also study the problems very closed to the FSTSP.

Recently, the routing problems with multiple drones have been studied. The Vehicle Routing Problem with Drones (VRPD) is first considered by Wang et al. \cite{wang2017}. Several theoretical aspects have been studied in terms of bounds and worst cases. An extension of that work is studied in \cite{poikonen2017}, where the authors consider more practical aspects such as drone endurance and cost. In addition, connections between VRPD and other classes of VRPs have been made in the form of bounds and asymptotic results. The extension to the FSTSP called The multiple flying sidekicks traveling salesman problem is presented in \cite{Murray2020} where a truck and multiple drones are synchronized to deliver parcels. A three-phase heuristic is proposed to solve instances with 100 customers and 4 drones. The $k$-Multi-visit Drone Routing Problem ($k$-MVDRP) in \cite{Poikonen2020} considers a tandem between a truck and $k$ drones. The new characteristic of the problem is that each drone is now capable of carrying multiple packages to deliver to customers. Unlike many papers in the literature, the $k$-MVDRP model also allows the specification of a drone energy drain function that takes into account the weight of parcels.

\begin{figure}[ht]
\centering
\includegraphics[scale = 0.28]{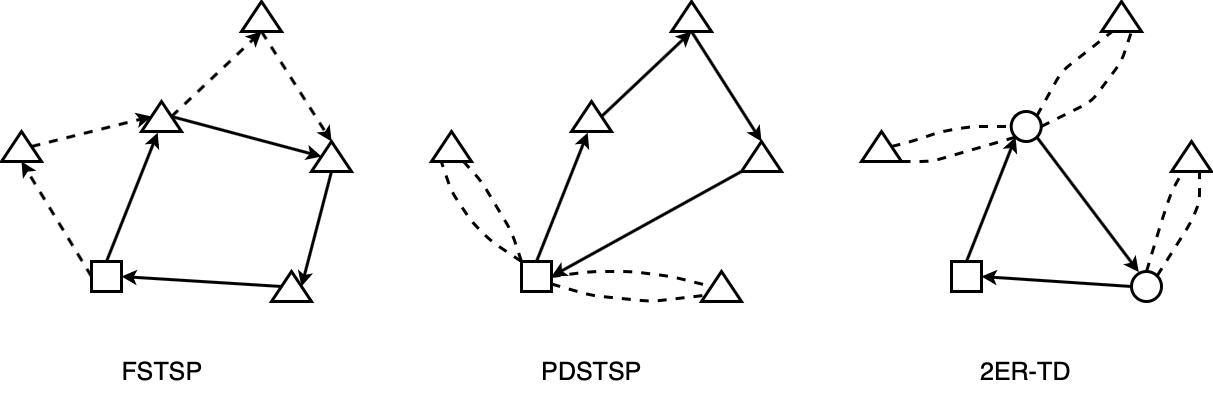}
\caption{Comparison among three problems: FSTSP, PDSTSP, and 2ER-TD}
\label{fig2}
\end{figure}

In \cite{murray2015flying}, the authors also study the second problem called the Parallel Drone Scheduling TSP (PDSTSP). In this problem, there is no truck-drone synchronization. Instead, a single truck is responsible for parcels far from the distribution centre (DC), and the drones are responsible for serving customers in its flight range around DC. In \cite{murray2015flying}, a formal definition, a mathematical programming formulation, and a simple heuristic solution approach based on greedy and local search strategies are provided. The problem is also studied in \cite{Saleu2018} in which the authors propose an iterative two‐step heuristic. The first step (or coding step) transforms a solution into a customer sequence, and the second step (also called decoding step) decomposes the customer sequence into a tour for the vehicle and trips for the drones. The decoding procedure is expressed as a bi-criteria shortest path problem and is carried out by dynamic programming. The main difference among the solutions of three problems FSTSP, PDSTSP, and 2ED-TD is illustrated in Figure \ref{fig2}. As specified in Figure \ref{fig1}, the dotted lines represent the drone trips while the solid ones show the truck trip.

Finally, we also note that in \cite{Kitjacharoenchai2019}, the authors consider a problem named the Two Echelon Vehicle Routing Problem with Drones (2EVRPD). But this problem is different from one considered in our paper in several aspects. First, the 2EVRPD is more similar to the FSTSP. In fact, the 2EVRPD extends the FSTSP by allowing multiple drones and multiple trucks to make deliveries with the consideration of both trucks’ and drones’ capacities. The drone is allowed to carry multiple packages to make multiple deliveries to different customers before merging with the truck. Another difference is that the 2EVRPD considers the objective function of minimizing the total delivery time, i.e. the total arrival time of both trucks and drones at the depot after completing the deliveries. Similarly to the 2EVRPD, the Two-Echelon Ground Vehicle and UAV cooperated Routing Problem (2E-GU-RP) introduced in \cite{Lou2017} is also defined differently from our problem. In contrast to the 2ER-TD, the drone has to launch and land on the truck frequently to change or charge its battery while the truck is moving on the road network. Besides, only one drone is also considered in the 2E-GU-RP.

\section{Mixed integer linear programming formulations}
\label{formu}
In this section, we propose mixed integer linear programming (MILP) formulations for two variants of the problem. Let $W_i$ be the set of customers that can be serviced from node $v_i$ ($v_i \in V$) and $V_k$ be the set of nodes from which drones can service customer $w_k$ ($w_k \in W$). Denote $v_n$ is a copy of the depot $v_0$. The formulation for the 2ER-TDs uses the following variables:

\begin{itemize}
    \item $y_i$: binary variables equal to 1 if truck visits node $v_i \in V$, to 0 otherwise.
    \item $x_{ij}$: binary variables equal to 1 if truck travels from node $v_i$ to node $v_j$ ($v_i, v_j \in V \cup \{v_n\}), i \neq j$).
    \item $z_{ik}$: binary variables equal to 1 if customer $w_k$ is serviced by a drone flies from node $v_i$ ($v_i \in V, w_k \in W_i$).
    \item $a_i$: positive real variable representing arrival time of truck to node $v_i \in V \cup \{v_n\}$. 
    \item $s_i$: positive real variable representing the duration the truck waits for the drones at node $v_i  \in V$.
\end{itemize}

The 2ER-TDs can be stated as:

\begin{align}
\textrm{Minimize} \, \qquad &a_n \label{eq:obj1}\\
\textrm{subject to} \qquad \;\; \sum_{v_i \in V}x_{0i} = 1; \; &\sum_{v_i \in V}x_{i0} = 0 \label{eq:c1}\\ \sum_{v_i \in V}x_{in} = 1; \; &\sum_{v_i \in V}x_{ni} = 0 \label{eq:c2}\\
\sum_{v_i \in V}x_{ij} &= y_j \qquad \forall v_j \in V \setminus \{v_0\}  \label{eq:c3}\\
%\sum_{j \in V}x_{ij} &= y_j \qquad \forall j \in V \setminus \{n\}  \label{eq:c4}\\
\sum_{v_j \in V} x_{ji} &= \sum_{j \in V}x_{ij} \qquad \forall v_i \in V \setminus \{v_0, v_n\} \label{eq:c5}\\
z_{ik} &\leq y_i \qquad \forall v_i \in V, w_k \in W_i \label{eq:c6}\\
\sum_{w_k \in W_i}z_{ik} &\geq y_i \qquad \forall v_i \in V\setminus \{v_0\} \label{eq:c7}\\
%z_{ik} &\leq f_{ik} \qquad \forall i \in V, k \in W \label{eq:c7}\\
\sum_{v_i \in V_k}z_{ik} &= 1 \qquad \forall w_k \in W \label{eq:c8}\\
\sum_{k \in W_i} z_{ik} &\leq u \qquad \forall v_i \in V \label{eq:c9}\\
%%z_{ik} &\leq f_{ik} \qquad \forall i \in V, \forall k \in W \label{eq:c10} \\
%\sum_{i \in V} \sum_{k \in W} z_{ik}&= m \qquad \label{eq:c11}\\
a_j \geq a_i + s_i + t_{ij}x_{ij} - M(1 - &x_{ij}) \qquad v_i \in V,  v_j \in V \cup \{v_n\}, j \neq 0 \label{eq:c12}\\
a_i &\leq My_i \qquad \forall v_i \in V \setminus \{v_0\} \label{eq:c12prime}\\
s_i &\geq \sum_{k \in W_i} 2t'_{ik}z_{ik} \qquad \forall v_i \in V \label{eq:c13}\\
x_{ij}& \in \{0,1\}  \qquad \forall v_i, v_j \in V \cup \{v_n\} \label{eq:c14}\\
y_i& \in \{0,1\}  \qquad \forall v_i \in V \label{eq:c15}\\
z_{ik} &\in \{0,1\} \qquad \forall v_i \in V \setminus \{v_0\}, w_k \in W_i\label{eq:c16}\\
a_i, s_i &\geq 0 \qquad \forall v_i \in V \cup \{v_n\} \label{eq:c17}.
\end{align}

The objective function (\ref{eq:obj1}) minimizes the time coming back to the depot of both truck and drones. Constraints (\ref{eq:c1}) require that the truck must start its route at the depot node $v_0$ and never come back there. Constraints (\ref{eq:c2}) mean the truck must return to and stop at the copy node of the depot $v_n$. Constraints (\ref{eq:c3}) imply that if the truck passes by a node $v_j$, it must travel on an arc going in $v_j$. Constraints (\ref{eq:c5}) ensure the flow conservation for the truck. Constraints (\ref{eq:c6}) show that if there is a customer serviced by a drone launched from $v_i$, the truck must pass by $v_i$. Constraints (\ref{eq:c7}) ensure that a truck node is visited only if at which at least one drone is launched to service a customer. An optimal solution never violates these constraints; they can thus be removed. However, we still add them to the model to avoid creating redundant visited truck nodes in solutions of MILP solvers on instances which are not solved to optimality. Constraints (\ref{eq:c8}) enforce every customer must be serviced while constraints (\ref{eq:c9}) make sure that there are at most $u$ drones launched from any node visited by the truck. Constraints (\ref{eq:c12}) show the relationship between the arrival times at nodes $v_i$ and $v_j$ if the truck travels from $v_i$ to $v_j$. Constraints (\ref{eq:c12prime}) imply that if a truck node is not visited, the arrival time at that node must be zero. These constraints are redundant but we still keep them to get rational values of $a$-variables. Here, $M$ in Constraints (\ref{eq:c12}) and (\ref{eq:c12prime})is a sufficiently large number and can be estimated by an upper bound of the objective value. Constraints (\ref{eq:c13}) compute the time that truck must wait for drones at each node $v_i \in V$. The remaining constraints define the variables' domain.

The formulation of the 2ER-TDm is more complex. Instead of using the two-index $z$-variables as in formulation above, we use three-index binary variables $z_{lik}$ representing if drone $l \in U$ flies from node $v_i \in V \setminus \{v_0\}$ to service customer $w_k \in W_i$. The 2ER-TDm can be modeled as:

\begin{align}
\textrm{subject to} \qquad \;\; (\ref{eq:obj1}), (\ref{eq:c1}), (\ref{eq:c2}), (\ref{eq:c3}), &(\ref{eq:c5}), (\ref{eq:c12}), (\ref{eq:c12prime}), (\ref{eq:c14}), (\ref{eq:c15}), (\ref{eq:c17}) \nonumber \\
z_{lik} &\leq y_i \qquad \forall l \in U, v_i \in V, w_k \in W_i \label{eq:c18}\\
\sum_{l \in U}\sum_{w_k \in W_i}z_{lik} &\geq y_i \qquad \forall v_i \in V\setminus \{v_0\} \label{eq:c18prime}\\
%x_{ijk} &\leq f_{jk} \qquad \forall j \in V, k \in W_j \label{eq:c8}\\
%z_{ijk} &\leq f_{jk} \qquad \forall i \in D, j \in V, k \in W \label{eq:c8}\\
\sum_{l \in U}\sum_{v_i \in V_k}z_{lik} &= 1 \qquad \forall w_k \in W \label{eq:c19}\\
%\sum_{i \in D} \sum_{j \in V} \sum_{k \in W} z_{ijk}&= m \qquad \label{eq:c10}\\
s_i &\geq \sum_{w_k \in W_i} 2t'_{ik} z_{lik} \qquad \forall v_i \in V, l \in U \label{eq:c20}\\
z_{lik} &\in \{0,1\} \qquad \forall l \in U, v_i \in V, w_k \in W_i\label{eq:c21}
\end{align}

Constraints (\ref{eq:c18}), (\ref{eq:c18prime}), (\ref{eq:c19}), and (\ref{eq:c20}) have the same meaning as constraints (\ref{eq:c6}), (\ref{eq:c7}), (\ref{eq:c8}), and (\ref{eq:c13}) in the formulation proposed for the 2ER-TDs, respectively.

The experiments show that our formulations can solve only small-scale problem instances within a reasonable CPU time. To fill this methodological gap and to solve larger instances that can arise in practice, we introduce a dedicated Greedy Randomized Adaptive Search Procedure in the next section.

\section{A metaheuristic}
\label{meta}

We now describe a Greedy Randomized Adaptive Search Procedure (GRASP) algorithm to find near-optimal solutions for the problems. This algorithm is selected due to its simplicity, speed, and efficacy. The GRASP was first introduced by \cite{Feo} and has been then applied to solve a number of VRPs such as \cite{Prins2009, ha2018min}. An overview of the heuristic can be found in Algorithm \ref{grasp-algorithm} where $f(S)$ represents the objective value of solution $S$. First, a greedy randomized heuristic \texttt{RandomizedHeuristic}() is used to provide a feasible solution. Then, this solution is improved by local search operators \texttt{LocalSearch}(). If the solution of the current iteration is better than the current best one, the latter is updated. The process is repeated until a given number of iterations $n_{max}$ is reached. We now describe in details two main components of the algorithm: \texttt{RandomizedHeuristic}() and \texttt{LocalSearch}().

\begin{algorithm}[H]
$f(S_{best}$) = $+\infty$\;
$iteration = 0$\;
\While {$iteration < n_{max}$} {
$iteration = iteration + 1$\;
$S$ = \texttt{RandomizedHeuristic}()\;
$S$ = \texttt{LocalSearch}(S)\;
\If {$f(S) < f(S_{best})$} {
$S_{best} = S$\;
}
}
return $S_{best}$\;
\caption{An overview of the GRASP heuristic} \label{grasp-algorithm} 
\end{algorithm}

\subsection{Greedy randomized heuristic}
Our greedy randomized heuristic \text uses the idea of cheapest insertion, i.e. we construct a solution by consecutively adding drone (or truck) nodes that provide the cheapest insertion cost until all the drone nodes are serviced. To add the randomness, we select an unvisited truck node or an unserviced drone node to insert into the partial solution based on the \textit{roulette wheel selection} algorithm ensuring that a cheaper insertion would be performed with higher probability. More precisely, the probability of performing the operation $i$ is computed as follows:
\begin{align}
    p_i = \frac{1/cost_i}{\sum_{j \in 
    O}1/cost_j} \label{ic}
\end{align}

where $cost_j$ is the cost of operation $j$ and $O$ is the set of possible operations. To define the next unserviced customer $w_k$ to be added to the current solution, we perform the following steps:
\begin{itemize}
    \item Step 1: for each unvisited truck node $v_j$ and each arc $(v_i, v_{i+1})$ in the current solution, we compute the insertion cost of $v_j$ between two nodes $v_i$ and $v_{i+1}$ as $IC_j = t_{ij} + t_{j(i+1)} - t_{i(i+1)}$. We then choose the insertion position for each unvisited truck node following the probability defined by formula (\ref{ic}). 
    \item Step 2: for each unserviced drone node $w_k$ and each truck node $v_i$ that can service $w_k$, the assignment cost $AC_{ki}$ is computed by $ AC_{ki} = WT^a_{i} - WT^b_{i} + IC_i$ if $v_i \in V$ is a visited truck node, or $AC_{ki} = WT^a_{i} - WT^b_{i}$  if $v_i \in V$ is not visited before. Here, $WT^a_{i}$ and $WT^b_{i}$ are the waiting times of the truck at node $v_i$ after and before the assignment of truck node $v_i$ to drone node $w_k$, respectively. In the case of single trip drone, $w_k$ is assigned to truck node $v_i$ iff there is at least one available drone that does not service any drone node at $v_i$. If there does not exit such a truck node, the current process is stopped and a new iteration of GRASP is started. The truck's waiting time at node $v_i$ after assigning $w_k$ to $v_i$ is computed as $WT^a_{i} = \max(WT^b_i, 2t'_{ik})$. For the problem with multiple trip drone, $w_k$ is assigned to the drone $l$ with the least flying time. In this case, the truck's waiting time at node $v_i$ is calculated as $WT^a_{i} = \max(WT^b_i, FT_l + 2t'_{ik})$ where $FT_l$ is the flying time of drone $l$. We then choose the assigned truck node for each unserviced drone node following the probability defined by formula (\ref{ic}). At the end of this step, we define the insertion cost $IC'_k$ of each unserviced drone node $w_k$ as $IC'_k = AC_{ki}$ where $v_i$ is the selected truck node to service $w_k$.
    \item Step 3: from the insertion costs of all unserviced drone nodes computed in the Step 2, we select a drone node to service in the current solution following the probability defined by formula (\ref{ic}).
\end{itemize}

%Đối với bài Single, số khách hàng bị giới hạn bởi số drone nên trước khi vào bước 2 ta tìm một cặp điểm ( khách hàng j, node i) với điều kiện khách hàng j  chỉ có thể phục vụ duy nhất bởi node i và node i chỉ còn duy nhất một suất drone. Sau đó chọn luôn cặp (j, i) này và quay lại bước 1. Nếu nghiệm hiện tại không thỏa mãn tức tồn tại một khách hàng không thể được phục vụ, ta quay lại thực hiện từ đầu sinh ra nghiệm mới.

The algorithm iteratively adds an unserviced drone node and repeats this procedure until all the drone nodes are serviced. At each iteration, the objective of using the roulette wheel selection in step 1 is to add the randomness to the decision related to sequencing truck nodes while that in step 2 is to handle the assignment of drone node and truck node. These combined with the least roulette wheel selection in step 3 allow our GRASP algorithm to diversify the search and explore wider solution spaces.

\subsection{Local search}
The 2ER-TDs includes three decision levels: the determination of nodes visited by the truck, the sequence among them, and the assignment of customers to the truck nodes. An additional decision layer that assigns drones to customers is considered in the 2ER-TDm. Our local search operators are designed in such a way that they can handle all these decision layers. 
\subsubsection{Local search for truck tour}
In both variants 2ER-TDs and 2ER-TDm, we first improve the truck tour using three classical local searches proposed for the TSP as follows: 
\begin{itemize}
    \item \textbf{relocateTruckNode}: relocate a truck node to other positions in the truck tour
    \item \textbf{swapTruckNode}: swap the position of two truck nodes in the tour
    \item \textbf{2-opt}: remove two edges from the tour, and reconnects the two created paths.
\end{itemize}
 We note that these moves do not impact the assignment of customers. They also do not remove any visited truck node neither add any unvisited node to the current solution. After the local search for the truck tour is performed, we compute and save the best position of inserting each unvisited truck node into the solution for later use.

\subsubsection{Local search for 2ER-TDs}
For the 2ER-TDs, at each truck node, a drone can perform only one flight and at most $u$ customers are serviced. The important decision in this case is to define the assignment of customers to truck nodes. Thus, the main objective of the local searches proposed for the 2ER-TDs is to re-assign the customers to other truck nodes:
\begin{itemize}
    \item \textbf{swapAssignment1}: we choose two customers and try to swap their assigned truck nodes. More precisely, let $assign(i, k)$ be an assignment of node $v_i$ to customer $w_k$. Given a feasible solution with two assignments $assign(i, k)$ and $assign(i', k')$, this local search tries to find a better solution by verifying assignments $assign(i, k')$ and $assign(i', k)$ which are feasible w.r.t the flying range constraint.
    \item \textbf{reAssignment1}: we try to reassign a customer to other truck nodes. Let $assign(i, k)$ be a current assignment of truck node $v_i$ to customer $w_k$. When we try to re-assign customer $w_k$ to a different truck node $v_j$ at which there is still an available drone, two cases must be taken into accounts. If $v_j$ is not visited in the current solution, it is added to its best position found at the end of the local search for a truck tour. The re-assignment cost must thus include this insertion cost of $v_j$. In the second case when $v_j$ is already visited, the re-assignment cost only depends on the additional waiting time of the truck at $v_j$ when serving customer $w_k$. After the move is performed, if $v_i$ does not service any customer, it is removed from the solution and its two adjacent nodes are directly linked in the new solution.
\end{itemize}

To speed up the local search, it is necessary to compute the variation of the objective function for each move in $\mathcal{O}(1)$. For this purpose, we use two vectors $t^f_i$ and $t^s_i$ to record the flying time of drones at each visited truck node $v_i$ to its assigned furthest and second furthest customers, respectively. Our local searches \texttt{swapAssignment1} and \texttt{reAssignment1} basically perform two different moves: add a customer to a truck node or remove a customer from a truck node. In the first case of adding customer $w_k$ to be serviced from truck node $v_i$, the increase of objective function between before and after the move can be computed in $\mathcal{O}(1)$ by $\delta = \max(0, 2t'_{ik} - t^f_i)$. If we remove customer $w_k$ from the service of node $v_i$, the decrease of objective value is $\delta = 0$ if $w_k$ is not the furthest customer serviced from $v_i$ (i.e., $2t'_{ik} < t^f_i$). In case $w_k$ is the current furthest customer serviced from $v_i$ ($2t'_{ik} = t^f_i$), the decrease is calculated by $\delta = t^f_i - t^s_i$. 
%Để có thể check cost trong O(1), tại mỗi node chúng ta cần tính và lưu lại khách hàng có thời gian bay xa nhất và xa nhì. Khi ảnh hưởng tới một node ta nhìn theo hai học độ chính là chèn một khách hàng vào hoặc bỏ một khách hàng ra khỏi node. Xet trường hợp chèn một khách hàng mới vào, cost mới của node được tính bằng so sánh hai lần khoảng cách khách hàng mới tới node và cost hiện tại node đó. Xét trường hợp còn lại là bỏ một khách hàng ra khỏi node, nếu khách hàng đó không phải là khách hàng có thời gian bay xa nhất trong danh sách phục vụ của node đang xét thì cost của node không bị ảnh hưởng. Nhưng nếu khách hàng bị bỏ ra khỏi node là khách hàng có khỏang cách xa nhất thì cost hiện tại bị ảnh hưởng và sẽ được thay thế nhanh bằng hai lần khoảng cách của khách hàng thứ nhì như đã nói trên  (HMH: Can noi ro hon.)

%Local search 2: 
%TH 1: Sau khi đổi, node phục vụ cũ không còn khách hàng thì bắt buộc cần phải xóa trong tour đi. Sau khi xóa, ta có đường đi trực tiếp giữa hai node liền kề với node cũ
%TH 2: Đổi sang node mới mà node đó chưa có trong tour truck dẫn đến cần phải chèn node mới vào tour. Check cost hay vị trí chèn node mới vào tour đều đã được tính trước như phần 3.1. Lưu ý nếu xảy ra hai trường hợp đồng thời mà vị trị chèn mới liền kề với node cũ thì ta thay node cũ thành node mới trong tour

\subsubsection{Local search for 2ER-TDm}
%Phức tạp hơn bài Single trip, Multiple trips cho phép drone có thể bay nhiều lần tại một node để phục vụ khách hàng. Như vậy số lượng khách hàng tại mỗi node không bị ảnh hưởng giới hạn bởi số drone nhưng các khách hàng trong một node lại có ảnh hưởng tới nhau. Hiểu sâu hơn, với mỗi node khi đã biết các khách hàng cần phải phục vụ bởi node đó thì việc tính được cost optimal của node đó cũng là một bài toán NP-hard, chính là bài toán Scheduling Jobs. Tiếp theo nếu ta thêm hoặc bỏ một khách hàng đi thì các khánh hàng còn lại cần phải tính lại để tìm drone nào sao cho đảm bảo tối ưu optimal tại mỗi node. Điều này dẫn đến không gian nghiệm bài toán lớn hơn bài Single cấp số mũ. Do đó cần nhiều loại local search hơn và có các local search giải quyết việc tối ưu bài toán trong từng node. 
%Để có thể check cost tăng trong O(1) tại mỗi node chúng ta cần tính và lưu lại các drone có tổng thời gian bay là lớn nhất, nhì và ba cùng với drone có tổng thời gian bé nhất. 
%3.3.1 
An additional decision layer that must be done in the case of the 2ER-TDm is to assign drones to customers. Denote \textit{droneAssignment($i, l, k$)} is the assignment decision representing that at truck node $v_i$, drone $l$ is launched to service customer $w_k$. We use the following local searches to re-optimize the drone-customer assignment at each truck node:
\begin{itemize}
    \item \textbf{relocateDroneAssignment1}: we choose a customer and try to assign it to a different drone of the same truck node. To be more precise, given an assignment \textit{droneAssignment($i, l, k$)} of the current solution, this operator investigates other drone assignments of type \textit{droneAssignment($i, l', k$)}. 
    
    \item \textbf{swapDroneAssignment1}: at a truck node $v_i$, we choose two customers $w_k$ and $w_{k'}$ currently assigned to two drones $l$ and $l'$, respectively. We then try to re-assign $w_k$ to drone $l'$ and $w_{k'}$ to drone $l$. 
    
    \item \textbf{relocateDroneAssignment2}: similarly to \texttt{relocateDroneAssign}-\texttt{ment1}, this local search tries to assign two customers instead of one to a different drone of the same truck node. 
    
    \item \textbf{swapDroneAssignment2}: this local search swaps two customers serviced by the same drone and one customer of another drone. Given assignments \textit{droneAssignment(i, l, k)}, \textit{droneAssignment(i, l, k')}, and \textit{droneAssignment(i, l', k")} of the current solution, this operator tries to find better solutions by verifying the neighborhoods with assignments of type \textit{droneAssignment(i, l', k)}, \textit{droneAssignment(i, l', k')}, and \textit{droneAssignment(i, l, k")}.
    
    \item \textbf{triangleDroneAssignment}: we choose three customers assigned to different drones and re-assign each customer to a different drone in the triangle fashion. More precisely, suppose we have three assignments \textit{droneAssignment(i, l, k)}, \textit{droneAssignment(i, l', k')}, and \textit{droneAssignment(i, l", k")}. This local search investigates the neighborhoods with assignments \textit{droneAssignment(i, l, k')}, \textit{droneAssignment(i, l', k")}, and \textit{droneAssignment(i, l", k)}.
\end{itemize}

%Tối ưu trong một node
%Trong mỗi node đã biết rõ danh sách các khách hàng cần phải phục vụ, ta thực hiện gồm bốn loại local search:
%Local search 1: Chọn 1 khách hàng sau đó đổi sang drone mới.
%Local search 2: Chọn 2 khách hàng sau đó đổi drone phục vụ cho nhau
%Local search 3: Chọn 2 khách hàng sau đó cùng đổi sang một drone mới
%Local search 4: Chọn 2 khách hàng cùng drone và 1 khách hàng drone khác sau đó đổi drone cho nhau
%Local search 5: Chọn 3 khách hàng khác drone và tráo đổi lần lượt cho nhau( 1->2, 2->3, 3->1)
%Ngoài local search ra, chúng tối bổ sung một thuật toán greedy tại mỗi node khi đã có danh sách các khách hàng như sau: Duyệt lần lượt các khách hàng có cost tăng dần, tại mỗi khách hàng j  sẽ được gán vào drone có tổng thời gian bay nhỏ nhất. Nếu cost của node sau bước greedy này nhỏ hơn cost hiện tại, ta thay thế vào nghiệm ban đầu.

After the local search operators above are terminated, we add a simple greedy heuristic for each truck node to get slightly better results. The customers assigned at a truck node are first sorted in a list of increasing the flying time between customers and the truck node. Then following this list, a customer is added to the drone of the least flying time. 

We also use \texttt{swapAssignment} and \texttt{reAssignment} proposed for the 2ER-TDs to change the customer-truck node assignment in the 2ER-TDm. The only difference is that, whenever a customer is re-assigned to a truck node, it is always added to the drone of the least flying time. Moreover, because the 2ER-TDm is more difficult than the 2ER-TDs, we use additional local searches as follows: 
%3.3.2 Tối ưu mảng covering:
%Khi ta thực hiện thêm, xóa,  hoặc sửa một khách hàng trong một node thì việc tính cost optimal của node đó trong O(1) gần như không thể vì là bài toán NP-Hard. Do đó, chúng tôi đã cố gắng tăng nhiều loại local search trong mỗi node nhất có thể để đảm bảo giảm tối đa nhất có thể sai lệch tính check cost khi ta thay đổi node phục vụ của một khách hàng. Ngoài ra còn vì số lượng khách hàng trong mỗi node sẽ nhỏ hơn. Đồng thời việc chèn một khách hàng vào một node sẽ được gán vào drone có tổng thời gian bay nhỏ nhất hiện tại của node đó. Đầu tiên chúng tôi gồm có:
%Local search 1: đổi node phục vụ của một khách hàng sang node mới 
%Local search 2: Đổi node phục vụ của 2 khách hàng sang cùng một node mới
%Local search 3: Chọn 2 khách hàng khác node phục vụ sau đó đổi node phục vụ cho nhau
%Local search 1 và 2 đều ảnh hưởng tới tour truck, ta xử lý tương tự như bài Single. Thêm vào đó như nói ở trên, khi chèn, xóa hay sửa khách hàng của một node chúng ta cần phải tìm optimal cost chính xác để tránh check cost sai khiến nghiệm hiện tại không thể nhảy đến một local optimal tốt được. Việc thực hiện lồng local search vào để check là điều không khả thi. Do đó chúng tôi bổ xung 1 local search tiêu biểu kết hợp với việc reduce lại cost node để check:
%Local search 4: Chọn 2 khách hàng cùng một node nhưng khách drone, đổi drone khách hàng thứ 2 sang drone khách hàng thứ nhất; sau đó đổi node phục vụ của khách hàng thứ nhất sang node mới.

\begin{itemize}
    \item \textbf{reAssignment2}: this operator works similarly to \texttt{reAssignment1}, but we try to re-assign two customers serviced from the same truck node to other truck nodes.

    \item \textbf{zigzagAssignment}: we choose two customers $w_k$ and $w_{k'}$ assigned to a truck drone $v_i$ and serviced by two different drones $l$ and $l'$. We then change the drone servicing the customer $w_{k'}$ to $l$ and re-assign the customer $w_k$ to another truck node $v_{i'}$. Suppose we have two assignments \textit{droneAssignment(i, l, k)} and \textit{droneAssignment(i, l', k')} in the current solution,  this operator searches for better solutions by checking assignments of type \textit{droneAssignment(i', l*, k)} and \textit{droneAssignment(i, l, k')}. Here, $l^*$ denotes the drone of the least flying time at truck node $v_{i'}$.
\end{itemize}

The local search operators and their order are described in the Algorithm \ref{algo:ls}. Whenever a local search or the greedy heuristic improves the current solution, the flag $\beta$ is set to true and the process is repeated. If there is no improvement generated, the procedure is stopped. In general, there are two fashions of implementing the local search: first improvement and best improvement. In this research, the first improvement is selected because it is faster. Finally, it is worth noting that, to compute the change of objective values of the solutions before and after local search moves in $\mathcal{O}(1)$ in the case of 2ER-TDm, we pre-compute and record at each visited truck node the drones of the largest, second largest, and least flying time.

\begin{algorithm}[!t]
  \textbf{Input:} a solution $S$, $variant$\; \Comment{$variant$ = 0 for single drone trip; = 1 for multiple drone trip}
  \textbf{Output:} a better solution $S$\;
  \While{$\beta =$ true}{
    \texttt{relocateTruckNode}$(S, \beta)$\;
    \texttt{swapTruckNode}$(S, \beta)$\;
    \texttt{2-opt}$(S, \beta)$\;
    \If{$variant = 0$}{
        \texttt{swapAssignment}$(S, \beta)$\;
        \texttt{reAssignment}$(S, \beta)$\;
    }
    \Else{
        \texttt{relocateDroneAssignment1}$(S, \beta)$\;
        \texttt{swapDroneAssignment1}$(S, \beta)$\;
        \texttt{relocateDroneAssignment2}$(S, \beta)$\;
        \texttt{swapDroneAssignment2}$(S, \beta)$\;
        \texttt{triangleDroneAssignment}$(S, \beta)$\;
        \texttt{greedyHeuristic}$(S, \beta)$\;
        \texttt{reAssignment1}$(S, \beta)$\;
        \texttt{reAssignment2}$(S, \beta)$\;
        \texttt{swapAssignment1}$(S, \beta)$\;
        \texttt{zigzagAssignment}$(S, \beta)$\;
    }
  }
  \Return $S$\;
  \caption{Local search procedure}
  \label{algo:ls}
\end{algorithm}

\section{Experimental results}
\label{result}
\subsection{Experimental setup}

Because there is no available benchmark instances for the problem, we create new instances with the parameters that simulate real-world scenarios as follows. All the nodes in the graph $n_{total} = n + m$ are randomly generated in a square of $d$ x $d$ dimension. First, the coordinates of $n$ truck nodes are created. The customers' coordinates are then generated such that each customer can be serviced from at least one truck node. In other words, whenever creating a customer that cannot be serviced from any truck node, we remove it and create others until the required number of feasible customers $m$ is reached. For all instances, the speeds of drone and truck are set to pre-defined values $\vartheta_d$ and $\vartheta_t$ in such a way that $\vartheta_d \geq \vartheta_t$. More precisely, the speed of the truck is 40~km/h while that of the drones is selected in five values: 40, 50, 60, 70, and 80~km/h. These values stay consistent with data used by recent works \cite{murray2015flying, Saleu2018}. The travel times between nodes are computed by $t_{ij} = \nicefrac{d_{ij}}{\vartheta_t}$ and $t'_{ik} = \nicefrac{d'_{ik}}{\vartheta_d}$ where $d_{ij}$ is the distance from a truck node $v_i \in V$ to another truck node $v_j \in V$, and $d'_{ik}$ is the distance from a truck node $v_i \in V$ to a drone node $w_k \in W$. Here, the distance $d_{ij}$ is calculated using Manhattan distance, while the distance $d'_{ik}$ is in Euclidean distance. The objective is to partially simulate the fact that the truck has to travel through a road network (which is longer) and the drone can fly directly from an origin to a destination. The drone has the flight endurance of 30 minutes as often used in other works \cite{murray2015flying, ha2018min, Saleu2018}. To avoid creating infeasible solutions in the case of single drone trip, we set the value of $u$ equal or greater than $u_{min}$, the minimum number of drones that must be used to service every customer while satisfying all constraints and the drone speed is set to the minimal value of 40~km/h. The value of $u_{min}$ is determined by solving the following MILP model:

\begin{align}
 u_{min} = \textrm{Minimize} \,\, &u \label{eq:obju}\\
\textrm{subject to} \qquad \;\; (\ref{eq:c8}), (\ref{eq:c9}), (\ref{eq:c16}) \nonumber 
\end{align}

The algorithms are implemented in C++ and run on an Intel(R) Core(TM) i5-9500 CPU @ 3.00GHz, 3000 Mhz, and 8GB RAM memory. CPLEX 12.10 is used whenever the MILP formulation needs to be solved. Different experiments have been carried out to evaluate the performance of the proposed methods and analyse the impact of parameters. All the instances and detailed results can be downloaded at \url{http://orlab.vn/home/download} or \url{http://orlab.com.vn/home/download}.

\subsection{Comparison of GRASP with MILP formulation}
In this section, to validate the MILP formulations, we report the results obtained by CPLEX in a running time for each instance limited to 1~hour. Another objective is to investigate the impact of data characteristics on the complexity of the problems. We also wish to observe the performance of the GRASP method when compared with the exact method.

The nodes in the graph are created in a square of dimension $d$ set to 20 and 30~km. Because CPLEX cannot solve to optimality large instances, we only consider small-size ones with  $n_{total} = \{20, 32\}$ and $n = \{\nicefrac{n_{total}}{4}, \nicefrac{n_{total}}{2}\}$. The number of drones is set to $u_{min}$, $u_{min}+1$, $u_{min}+2$, and $u_{min}+3$. Our instances are named $d-n-m$, where $d$ is the dimension of the square which includes the graph, $n$ is the number of truck nodes, and $m$ is the number of customers. The big $M$ in the MILP formulations is set to the objective value of solutions provided by the GRASP. And finally, the number of iterations $n_{max}$ of the GRASP is set to 5000. 

The obtained results of the algorithms in the cases $d=20$ and $d=30$ are reported in Tables \ref{d20} and \ref{d30}, respectively. Here, the meaning of column headings is as follows:
\begin{itemize}
    \item Data: the name of instances;
    \item $\vartheta_d$: the speed of drones;
    \item Obj: the objective value of solutions found by the methods;
    \item Time: the running time of the algorithms to find the final solutions (in seconds);
\end{itemize}
The instances that CPLEX cannot solve to optimality are marked with (-) while optimal solutions are marked with (*). Better solutions provided by the two methods are highlighted in bold.

\begin{table}[]
\resizebox{\textwidth}{0.70\columnwidth}{%
\begin{tabular}{|c|c|c|c|c|c|c|c|c|c|}
\hline
\multirow{3}{*}{Data} & \multirow{3}{*}{$\vartheta_d$} & \multicolumn{4}{c|}{2ER-TDs}                           & \multicolumn{4}{c|}{2ER-TDm}                           \\ \cline{3-10} 
                      &                           & \multicolumn{2}{c|}{MILP} & \multicolumn{2}{c|}{GRASP} & \multicolumn{2}{c|}{MILP} & \multicolumn{2}{c|}{GRASP} \\ \cline{3-10} 
                      &                           & Time         & Obj        & Time          & Obj        & Time         & Obj        & Time          & Obj        \\ \hline
\multicolumn{10}{|c|}{$u = u_{min}$}                                                                                                                                   \\ \hline
20-5-15&40&0.02&2.816*&0.03&2.816&0.05&2.433*&0.24&2.433\\ \hline
20-5-15&50&0.02&2.553*&0.04&2.553&0.04&2.187*&0.26&2.187\\ \hline
20-5-15&60&0.02&2.378*&0.05&2.378&0.09&1.250*&0.42&1.250\\ \hline
20-5-15&70&0.03&2.252*&0.06&2.252&0.06&1.093*&0.41&1.093\\ \hline
20-5-15&80&0.06&2.158*&0.07&2.158&0.06&0.975*&0.42&0.975\\ \hline
20-10-10&40&28.93&2.786*&0.07&2.786&11.44&2.206*&0.09&2.206\\ \hline
20-10-10&50&91.66&2.518*&0.07&2.518&8.13&1.845*&0.11&1.845\\ \hline
20-10-10&60&121.92&2.340*&0.08&2.340&2.65&1.492*&0.12&1.492\\ \hline
20-10-10&70&75.73&2.213*&0.08&2.213&1.18&1.300*&0.14&1.300\\ \hline
20-10-10&80&255.72&2.098*&0.07&2.098&0.85&1.156*&0.13&1.156\\ \hline
\multicolumn{10}{|c|}{$u = u_{min}$ + 1}    \\ \hline
20-5-15&40&0.01&2.738*&0.04&2.738&0.05&2.338*&0.25&2.338\\ \hline
20-5-15&50&0.02&2.425*&0.04&2.425&0.05&2.111*&0.30&2.111\\ \hline
20-5-15&60&0.04&2.221*&0.06&2.221&0.04&1.038*&0.47&1.038\\ \hline
20-5-15&70&0.05&2.075*&0.07&2.075&0.06&0.911*&0.46&0.911\\ \hline
20-5-15&80&0.20&1.966*&0.07&1.966&0.04&0.816*&0.46&0.816\\ \hline
20-10-10&40&14.18&2.392*&0.06&2.392&4.08&1.640*&0.13&1.640\\ \hline
20-10-10&50&19.40&1.969*&0.07&1.969&0.70&1.392*&0.15&1.392\\ \hline
20-10-10&60&13.62&1.766*&0.07&1.766&0.34&1.047*&0.16&1.047\\ \hline
20-10-10&70&7.82&1.621*&0.07&1.621&0.21&0.919*&0.17&0.919\\ \hline
20-10-10&80&5.79&1.512*&0.07&1.512&0.15&0.823*&0.17&0.823\\ \hline
\multicolumn{10}{|c|}{$u = u_{min}$ + 2}    \\ \hline
20-5-15&40&0.01&2.620*&0.04&2.620&0.04&2.227*&0.24&2.227\\ \hline
20-5-15&50&0.01&2.270*&0.04&2.270&0.05&2.022*&0.29&2.022\\ \hline
20-5-15&60&0.04&2.071*&0.06&2.071&0.05&0.892*&0.50&0.892\\ \hline
20-5-15&70&0.04&1.947*&0.07&1.947&0.03&0.786*&0.49&0.786\\ \hline
20-5-15&80&0.05&1.853*&0.08&1.853&0.03&0.707*&0.49&0.707\\ \hline
20-10-10&40&1.89&1.824*&0.05&1.824&1.72&1.434*&0.14&1.434\\ \hline
20-10-10&50&3.11&1.589*&0.06&1.589&0.27&1.146*&0.16&1.146\\ \hline
20-10-10&60&2.42&1.433*&0.06&1.433&0.12&0.834*&0.18&0.834\\ \hline
20-10-10&70&0.53&1.321*&0.07&1.321&0.10&0.736*&0.19&0.736\\ \hline
20-10-10&80&0.53&1.237*&0.06&1.237&0.05&0.663*&0.18&0.663\\ \hline
\multicolumn{10}{|c|}{$u = u_{min}$ + 3}    \\ \hline
20-5-15&40&0.01&2.290*&0.04&2.290&0.06&2.190*&0.24&2.190\\ \hline
20-5-15&50&0.02&2.072*&0.04&2.072&0.06&1.992*&0.29&1.992\\ \hline
20-5-15&60&0.04&1.927*&0.06&1.927&0.06&0.822*&0.43&0.822\\ \hline
20-5-15&70&0.05&1.823*&0.07&1.823&0.04&0.726*&0.44&0.726\\ \hline
20-5-15&80&0.05&1.745*&0.07&1.745&0.03&0.654*&0.45&0.654\\ \hline
20-10-10&40&0.26&1.319*&0.05&1.319&1.45&1.319*&0.15&1.319\\ \hline
20-10-10&50&0.20&1.135*&0.05&1.135&0.19&1.037*&0.16&1.037\\ \hline
20-10-10&60&0.17&1.013*&0.06&1.013&0.08&0.705*&0.19&0.705\\ \hline
20-10-10&70&0.12&0.925*&0.06&0.925&0.05&0.626*&0.19&0.626\\ \hline
20-10-10&80&0.12&0.859*&0.05&0.859&0.03&0.566*&0.19&0.566\\ \hline
\end{tabular}%
}
\caption{Comparison between MILP and GRASP with $d$ = 20}
\label{d20}
\end{table}

% \usepackage{multirow}
% \usepackage{graphicx}
\begin{table}[]
\resizebox{\textwidth}{0.70\columnwidth}{%
\begin{tabular}{|c|c|c|c|c|c|c|c|c|c|}
\hline
\multirow{3}{*}{Data} & \multirow{3}{*}{$\vartheta_d$} & \multicolumn{4}{c|}{2ER-TDs}                           & \multicolumn{4}{c|}{2ER-TDm}                           \\ \cline{3-10} 
                      &                           & \multicolumn{2}{c|}{MILP} & \multicolumn{2}{c|}{GRASP} & \multicolumn{2}{c|}{MILP} & \multicolumn{2}{c|}{GRASP} \\ \cline{3-10} 
                      &                           & Time         & Obj        & Time          & Obj        & Time         & Obj        & Time          & Obj        \\ \hline
\multicolumn{10}{|c|}{$u = u_{min}$}                                                                                                                                   \\ \hline
30-8-24&40&0.35&4.292*&0.09&4.292&3.38&4.145*&0.57&4.145\\ \hline
30-8-24&50&0.74&3.844*&0.10&3.844&6.54&3.726*&0.61&3.726\\ \hline
30-8-24&60&0.94&3.545*&0.15&3.545&3.98&3.111*&0.76&3.111\\ \hline
30-8-24&70&5.41&3.331*&0.19&3.331&8.03&2.613*&1.05&2.613\\ \hline
30-8-24&80&2.07&3.171*&0.21&3.171&9.20&2.442*&1.17&2.442\\ \hline
30-16-16&40&3600.03&-&0.15&4.490&3600.04&-&0.20&\textbf{3.983}\\ \hline
30-16-16&50&3600.84&-&0.18&3.977&3600.03&-&0.22&\textbf{3.556}\\ \hline
30-16-16&60&3600.02&-&0.23&3.672&3601.06&-&0.26&\textbf{3.254}\\ \hline
30-16-16&70&3600.03&-&0.24&3.455&3600.03&-&0.30&\textbf{2.975}\\ \hline
30-16-16&80&3600.02&-&0.23&3.292&3600.02&-&0.32&\textbf{2.766}\\ \hline
\multicolumn{10}{|c|}{$u = u_{min}$ + 1}     \\ \hline
30-8-24&40&0.27&4.134*&0.09&4.134&3.30&4.031*&0.54&4.031\\ \hline
30-8-24&50&0.39&3.717*&0.14&3.717&5.86&3.635*&0.59&3.635\\ \hline
30-8-24&60&0.58&3.300*&0.17&3.300&3.33&2.988*&0.74&2.988\\ \hline
30-8-24&70&3.03&3.086*&0.21&3.086&1.38&2.426*&1.06&2.426\\ \hline
30-8-24&80&2.81&2.925*&0.22&2.925&3.84&2.260*&1.19&2.260\\ \hline
30-16-16&40&3600.03&-&0.14&3.728&3600.64&-&0.23&\textbf{3.395}\\ \hline
30-16-16&50&3600.66&-&0.16&3.352&3600.02&-&0.27&\textbf{3.086}\\ \hline
30-16-16&60&3600.02&-&0.20&3.102&3600.02&-&0.34&\textbf{2.698}\\ \hline
30-16-16&70&3600.02&-&0.23&2.923&3600.12&-&0.38&\textbf{2.499}\\ \hline
30-16-16&80&3600.01&-&0.23&2.789&3600.02&-&0.41&\textbf{2.311}\\ \hline
\multicolumn{10}{|c|}{$u = u_{min}$ + 2}    \\ \hline
30-8-24&40&0.45&4.079*&0.08&4.079&1.72&3.878*&0.52&3.878\\ \hline
30-8-24&50&1.29&\textbf{3.669*}&0.13&3.673&4.90&3.512*&0.60&3.512\\ \hline
30-8-24&60&0.44&3.107*&0.17&3.107&2.57&2.933*&0.70&2.933\\ \hline
30-8-24&70&1.28&2.920*&0.19&2.920&1.51&2.311*&1.01&2.311\\ \hline
30-8-24&80&0.93&2.780*&0.20&2.780&1.04&2.145*&1.16&2.145\\ \hline
30-16-16&40&3600.01&-&0.13&\textbf{3.537}&3600.02&-&0.25&\textbf{3.382}\\ \hline
30-16-16&50&3600.01&-&0.15&\textbf{3.212}&3600.02&-&0.28&\textbf{3.030}\\ \hline
30-16-16&60&3600.02&-&0.17&\textbf{2.639}&3600.02&-&0.37&\textbf{2.470}\\ \hline
30-16-16&70&3600.03&-&0.21&\textbf{2.448}&3600.03&-&0.42&\textbf{2.302}\\ \hline
30-16-16&80&3600.03&-&0.21&\textbf{2.304}&3600.03&-&0.45&\textbf{2.112}\\ \hline
\multicolumn{10}{|c|}{$u = u_{min}$ + 3}    \\ \hline
30-8-24&40&0.67&\textbf{3.836*}&0.09&3.979&1.59&3.825*&0.53&3.825\\ \hline
30-8-24&50&0.94&3.478*&0.13&3.478&3.04&3.470*&0.60&3.470\\ \hline
30-8-24&60&1.53&3.033*&0.17&3.033&2.29&2.928*&0.68&2.928\\ \hline
30-8-24&70&0.96&2.615*&0.19&2.615&1.49&2.195*&0.96&2.195\\ \hline
30-8-24&80&0.92&2.470*&0.19&2.470&0.70&2.077*&1.09&2.077\\ \hline
30-16-16&40&3600.02&3.481&0.14&3.481&3600.02&-&0.24&\textbf{3.382}\\ \hline
30-16-16&50&3600.06&-&0.15&\textbf{3.023}&3600.01&-&0.28&\textbf{3.023}\\ \hline
30-16-16&60&3600.02&-&0.17&\textbf{2.530}&3600.03&-&0.37&\textbf{2.267}\\ \hline
30-16-16&70&3600.02&-&0.21&\textbf{2.354}&3600.02&-&0.42&\textbf{2.122}\\ \hline
30-16-16&80&3600.01&-&0.18&\textbf{2.223}&3600.04&-&0.46&\textbf{2.013}\\ \hline
\end{tabular}%
}
\caption{Comparison between MILP and GRASP with $d$ = 30}
\label{d30}
\end{table}

As can be seen in the result tables, the MILP-based methods can successfully solve to optimality all the instances but ones with 16 customers and 16 truck nodes. When the number of truck nodes increases, the problems become harder as expected. However, the hardness of the problems tends to be insensitive to the number of customers. This is similar to the covering tour problem studied in \cite{ha2012}, although compared with the covering tour problem, the 2ER-TD problems involve an additional decision layer related to the solution of the parallel machine scheduling problem at each visited truck node. Another interesting observation is that, although being more complex, there is no clear indication that shows 2ER-TDm is harder than 2ER-TDs for CPLEX. On instances with $n = \nicefrac{n_{total}}{2}$, CPLEX takes significantly more time to solve the 2ER-TDs formulation. Using more drones tends to make the problem harder to be solved to optimality, especially on 2ER-TDs problems. The impact of the drone speed on the problems' complexity is not clear. Increasing the drone speed does not lead to a longer or shorter running time of CPLEX in a steady fashion. 

The results also show a good performance of the GRASP on the tested instances. It finds all the optimal solutions provided by CPLEX but two instances 30-8-24 ($\vartheta_d = 50$) and 30-8-24 ($\vartheta_d = 40$). On open instances, i.e. those that CPLEX cannot solve to optimality, CPLEX cannot find any solution. The only exception is on the instance 30-16-16 ($\vartheta = 40$) when CPLEX and GRASP generate the same solution. It is worth mentioning that because we set $M$ in the two formulations to the objective values of solutions given by GRASP, the MILP-based methods are enforced to find solutions that are at least as good as those of GRASP. In other words, this corresponds to provide CPLEX an implicit upper bound before its execution. The running time of GRASP is quite fast, never exceeding 1.5 seconds. GRASP takes more time to handle 2ER-TDm because it includes more local search operators in this case.

As expected, the completion time coming back to the depot of all vehicles is reduced when the number of drones increases or the speed of drones is higher. Further computations show that increasing the drone speed by an amount of 10~km/h to 40~km/h can help the systems reduce the completion time from 9.19~\% to 36.79~\% in the case of 2ER-TDs and from 9.04~\% to 70.14~\% in the case of 2ER-TDm. The corresponding numbers for the context of increasing the number of drones by a quantity of one to three are 2.77~\%, 59.06~\%, 2.44~\%, and 52.75~\%, respectively.

\subsection{Comparison of 2ER-TD transportation mode with TSP mode}
\label{compareTSP}
In this section, we carry out the experiment to answer the question: ``Could the 2-echelon truck-drone combination increase the service quality when being applied in the delivery of commercial products?". Answering this question is not trivial. Drones can fly faster in a straight line while truck is often slower and has to follow the city block distance. However, drones must perform back-and-forth trips whenever they service a customer. We answer the question empirically by comparing the new transportation mode (2ER-TD mode) with the common one using truck only (TSP mode) in terms of the time coming back to the depot of all the vehicles. As such, 2ER-TD solutions are compared with optimal TSP tours.

We generate graphs in a square of dimension $d$ set to 30 and 40~km. The total number of nodes $n_{total}$ is set to 75 and 100. In this experiment, we suppose all the nodes in the graph are truck nodes, and also drone nodes. In other words, the truck nodes coincide with the customers. In the TSP mode, the trucks must pass through every customer while in the 2ER-TD mode, the truck must visit a customer location to launch the drones servicing that customer itself and the others. The number of drones is set to 2, 3, 4, and 5. The instances of this experiment are named $d-n_{total}$. For example, \texttt{40-100} represents an instance with 100 nodes generated in a square of dimension 40x40. Because the MILP-based method does not handle these large instances efficiently, we use the GRASP to generate 2ER-TD solutions for the comparison. The number of iterations $n_{max}$ of GRASP is set to 50,000 to more efficiently deal with larger instances. Finally, we permit drones to perform multiple trips in this experiment and optimal TSP tours are obtained with the state-of-the-art Concorde solver \cite{Applegate2006}. 

\begin{figure}[ht]
\centering
\includegraphics[scale = 0.56]{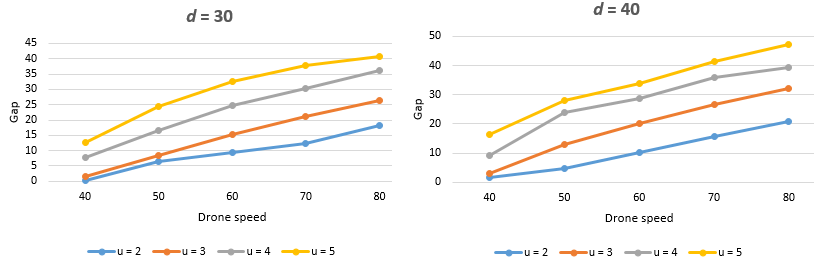}
\caption{Compare the two transportation modes when $n_{total} = 75$}
\label{TSPCompare75}
\end{figure}

\begin{figure}[ht]
\centering
\includegraphics[scale = 0.51]{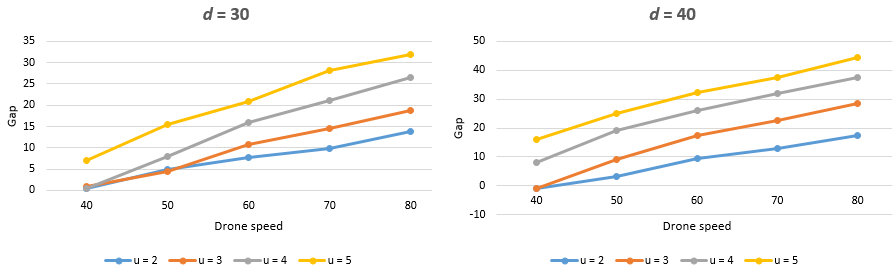}
\caption{Compare the two transportation modes when $n_{total} = 100$}
\label{TSPCompare100}
\end{figure}

The comparison results are reported in the Appendix and summarized in four graphs of Figures \ref{TSPCompare75} and \ref{TSPCompare100}. Each graph displays the relative gap (in percentage) calculated as the amount of time saved by using the 2ER-TD mode divided by the objective value of the corresponding optimal TSP solution. As expected, the higher the number of drones and their speed are, the larger the gap's values are. In particular, the gap depends on the drone speed in an approximately linear relationship. With the smallest value of the drone speed, the 2ER-TD can reduce up to nearly 20\% amount of time. The gaps increase even up to nearly 50\% when the speed of drones is doubled to 80~km/h. This clearly demonstrates the benefit of using the 2ER-TD in commercial deliveries. Another interesting observation is that the time reduction given by the 2ED-TD tends to be higher in regions with low customer densities. For a given setting in which the number of drones and their speed are fixed, the gap in the case $d = 40$ is, in general, greater than that in the case $d = 30$. The large distances between customers' locations bring advantages for using drones as a means of delivery.

\section{Conclusion}
\label{conclu}

In this article, we introduced two variants of the well-known two-echelon vehicle routing problem in which a truck works on the first echelon to transport parcels to intermediate depots while in the second echelon, drones are used to deliver parcels to customers. We proposed MILP programs and a GRASP-based metaheuristic with problem-tailored construction solutions and local search operators. Our local search moves can cover all the decision layers of the problems and run in $\mathcal{O}(1)$ using other data structures. Obtained results show the validity of our models and the performance of our metaheuristic on the tested instances. The experiments also demonstrate the benefit of using the 2ER-TD transportation mode in commercial delivery. 

The research perspectives are numerous. First, a clear limit in the evaluation of our methods is the lack of both upper and lower bounds (except for small instances for which optimal solutions are known). Developing an efficient exact solution method, such as branch-and-cut or branch-and-price, or finding tight lower bounds is certainly an important direction to follow. Exact solution algorithms have been developed for other types of routing problems where only a subset of nodes is selected to visit (e.g. the vehicle routing problem with service level constraints \cite{Bu2018}). The valid inequalities and separation algorithms used in these algorithms could certainly be adapted to our problems. Other more complex meta-heuristics (for example, the hybrid genetic algorithm in \cite{Minh2020}) could be also developed to investigate the performance of our GRASP on larger instances.

Finally, more complex and general variants of the problems would certainly deserve to be studied in future works. Especially, a strong limit in our problems is to use a single truck. Delivery service can typically involve multiple trucks. Extending the problem to this setting would definitely be an interesting research direction to follow.

%\section*{References:}
\bibliography{refer}{}
\bibliographystyle{plain}
\section*{Appendix}
The following table reports the detailed results for the experiment in Subsection \ref{compareTSP}. The column headings are as follows:
\begin{itemize}
    \item Data: the name of instances;
    \item $v_t, v_d$: the speed of truck and drones, respectively;
    \item $u$: the number of drones;
    \item TSP: the objective value of optimal TSP solutions;
    \item $|V_t|$: the number of truck nodes that are visited in the solution;
    \item Obj: the objective value of solutions found by GRASP;
    \item $T_d$: the total time that truck has to wait for drones;
    \item $T_t$: the travel time of truck without idle time that truck has to wait for drones;
    \item Gap: the relative gap (in percentage) computed as $Gap = 100\frac{TSP-Obj}{TSP}$
    \item Time: the running time in seconds of GRASP;
\end{itemize}

 \begin{table}[]
 \centering
\resizebox{0.7\textwidth}{0.75\columnwidth}{%
\begin{tabular}{|c|c|c|c|c|c|c|c|c|c|c|}
\hline
\multirow{2}{*}{\textbf{Data}} & \multirow{2}{*}{\textbf{$\vartheta_t$}} & \multirow{2}{*}{$\vartheta_d$} & \multirow{2}{*}{\textbf{$u$}} & \multirow{2}{*}{\textbf{TSP}} & \multicolumn{6}{c|}{\textbf{GRASP}}                                                         \\ \cline{6-11} 
                                   &                              &                              &                                  &                               & \textbf{$|V_t|$} & \textbf{Obj}   & \textbf{$T_d$} & \textbf{$T_t$} & \textbf{Gap} & \textbf{Time} \\ \hline
30-100 & 40 & 40   & 2                & 7.200           & 97        & 7.163  & 0.363 & 6.800  & 0.51  & 270.48 \\
30-100 & 40 & 40   & 3                & 7.200           & 95        & 7.133  & 0.433 & 6.700  & 0.93  & 267.15 \\
30-100 & 40 & 40   & 4                & 7.200           & 33        & 7.164  & 3.064 & 4.100  & 0.50  & 273.84 \\
30-100 & 40 & 40   & 5                & 7.200           & 41        & 6.702  & 2.402 & 4.300  & 6.92  & 274.45 \\
30-100 & 40 & 50   & 2                & 7.200           & 83        & 6.854  & 1.004 & 5.850  & 4.81  & 371.11 \\
30-100 & 40 & 50   & 3                & 7.200           & 79        & 6.882  & 1.032 & 5.850  & 4.42  & 370.65 \\
30-100 & 40 & 50   & 4                & 7.200           & 23        & 6.620  & 3.470 & 3.150  & 8.06  & 378.77 \\
30-100 & 40 & 50   & 5                & 7.200           & 21        & 6.094  & 2.794 & 3.300  & 15.36 & 380.45 \\
30-100 & 40 & 60   & 2                & 7.200           & 69        & 6.651  & 1.601 & 5.050  & 7.63  & 461.21 \\
30-100 & 40 & 60   & 3                & 7.200           & 67        & 6.430  & 1.480 & 4.950  & 10.69 & 469.16 \\
30-100 & 40 & 60   & 4                & 7.200           & 19        & 6.055  & 3.105 & 2.950  & 15.90 & 484.06 \\
30-100 & 40 & 60   & 5                & 7.200           & 18        & 5.701  & 2.751 & 2.950  & 20.82 & 487.00 \\
30-100 & 40 & 70   & 2                & 7.200           & 62        & 6.499  & 1.499 & 5.000  & 9.74  & 549.66 \\
30-100 & 40 & 70   & 3                & 7.200           & 23        & 6.156  & 2.606 & 3.550  & 14.50 & 558.94 \\
30-100 & 40 & 70   & 4                & 7.200           & 18        & 5.673  & 2.423 & 3.250  & 21.21 & 586.02 \\
30-100 & 40 & 70   & 5                & 7.200           & 17        & 5.167  & 2.367 & 2.800  & 28.24 & 589.85 \\
30-100 & 40 & 80   & 2                & 7.200           & 62        & 6.208  & 1.558 & 4.650  & 13.78 & 603.23 \\
30-100 & 40 & 80   & 3                & 7.200           & 24        & 5.855  & 2.805 & 3.050  & 18.68 & 614.52 \\
30-100 & 40 & 80   & 4                & 7.200           & 18        & 5.284  & 2.284 & 3.000  & 26.61 & 651.89 \\
30-100 & 40 & 80   & 5                & 7.200           & 15        & 4.897  & 2.247 & 2.650  & 31.99 & 656.30 \\
30-75  & 40 & 40   & 2                & 6.800  & 70        & 6.824  & 0.424 & 6.400  & -0.35 & 118.86 \\
30-75  & 40 & 40   & 3                & 6.800           & 71        & 6.704  & 0.404 & 6.300  & 1.41  & 118.50 \\
30-75  & 40 & 40   & 4                & 6.800           & 26        & 6.281  & 2.731 & 3.550  & 7.63  & 125.15 \\
30-75  & 40 & 40   & 5                & 6.800           & 27        & 5.930  & 2.380 & 3.550  & 12.79 & 126.44 \\
30-75  & 40 & 50   & 2                & 6.800           & 61        & 6.369  & 1.019 & 5.350  & 6.34  & 153.26 \\
30-75  & 40 & 50   & 3                & 6.800           & 19        & 6.218  & 3.168 & 3.050  & 8.56  & 156.12 \\
30-75  & 40 & 50   & 4                & 6.800           & 21        & 5.670  & 2.670 & 3.000  & 16.62 & 165.12 \\
30-75  & 40 & 50   & 5                & 6.800           & 17        & 5.137  & 3.037 & 2.100  & 24.46 & 166.55 \\
30-75  & 40 & 60   & 2                & 6.800           & 50        & 6.172  & 1.422 & 4.750  & 9.24  & 193.29 \\
30-75  & 40 & 60   & 3                & 6.800           & 22        & 5.762  & 2.412 & 3.350  & 15.26 & 199.44 \\
30-75  & 40 & 60   & 4                & 6.800           & 16        & 5.117  & 2.617 & 2.500  & 24.75 & 211.50 \\
30-75  & 40 & 60   & 5                & 6.800           & 16        & 4.580  & 2.380 & 2.200  & 32.65 & 212.64 \\
30-75  & 40 & 70   & 2                & 6.800           & 22        & 5.956  & 2.706 & 3.250  & 12.41 & 226.59 \\
30-75  & 40 & 70   & 3                & 6.800           & 24        & 5.372  & 2.372 & 3.000  & 21.00 & 236.96 \\
30-75  & 40 & 70   & 4                & 6.800           & 15        & 4.743  & 2.693 & 2.050  & 30.25 & 246.02 \\
30-75  & 40 & 70   & 5                & 6.800           & 10        & 4.228  & 2.028 & 2.200  & 37.82 & 246.02 \\
30-75  & 40 & 80   & 2                & 6.800           & 23        & 5.572  & 2.372 & 3.200  & 18.06 & 252.89 \\
30-75  & 40 & 80   & 3                & 6.800           & 14        & 5.006  & 2.856 & 2.150  & 26.38 & 268.40 \\
30-75  & 40 & 80   & 4                & 6.800           & 12        & 4.337  & 2.387 & 1.950  & 36.22 & 276.21 \\
30-75  & 40 & 80   & 5                & 6.800           & 9         & 4.026  & 1.826 & 2.200  & 40.79 & 277.72 \\
40-100 & 40 & 40   & 2                & 10.150 & 99        & 10.262 & 0.212 & 10.050 & -1.10 & 204.39 \\
40-100 & 40 & 40   & 3                & 10.150 & 40        & 10.241 & 4.641 & 5.600  & -0.90 & 205.78 \\
40-100 & 40 & 40   & 4                & 10.150          & 28        & 9.335  & 4.735 & 4.600  & 8.03  & 210.20 \\
40-100 & 40 & 40   & 5                & 10.150          & 33        & 8.515  & 3.765 & 4.750  & 16.11 & 211.60 \\
40-100 & 40 & 50   & 2                & 10.150          & 73        & 9.815  & 2.115 & 7.700  & 3.30  & 271.51 \\
40-100 & 40 & 50   & 3                & 10.150          & 32        & 9.218  & 4.468 & 4.750  & 9.18  & 276.38 \\
40-100 & 40 & 50   & 4                & 10.150          & 19        & 8.191  & 4.591 & 3.600  & 19.30 & 283.04 \\
40-100 & 40 & 50   & 5                & 10.150          & 20        & 7.617  & 3.617 & 4.000  & 24.96 & 284.72 \\
40-100 & 40 & 60   & 2                & 10.150          & 33        & 9.210  & 4.210 & 5.000  & 9.26  & 348.32 \\
40-100 & 40 & 60   & 3                & 10.150          & 28        & 8.373  & 4.073 & 4.300  & 17.51 & 358.24 \\
40-100 & 40 & 60   & 4                & 10.150          & 21        & 7.496  & 3.496 & 4.000  & 26.15 & 367.10 \\
40-100 & 40 & 60   & 5                & 10.150          & 14        & 6.878  & 3.528 & 3.350  & 32.24 & 369.04 \\
40-100 & 40 & 70   & 2                & 10.150          & 32        & 8.823  & 3.723 & 5.100  & 13.07 & 435.22 \\
40-100 & 40 & 70   & 3                & 10.150          & 22        & 7.841  & 4.241 & 3.600  & 22.75 & 452.30 \\
40-100 & 40 & 70   & 4                & 10.150          & 16        & 6.894  & 3.594 & 3.300  & 32.08 & 463.97 \\
40-100 & 40 & 70   & 5                & 10.150          & 16        & 6.338  & 3.338 & 3.000  & 37.56 & 466.11 \\
40-100 & 40 & 80   & 2                & 10.150          & 50        & 8.370  & 2.720 & 5.650  & 17.54 & 505.54 \\
40-100 & 40 & 80   & 3                & 10.150          & 20        & 7.244  & 3.694 & 3.550  & 28.63 & 531.87 \\
40-100 & 40 & 80   & 4                & 10.150          & 15        & 6.327  & 3.177 & 3.150  & 37.67 & 546.38 \\
40-100 & 40 & 80   & 5                & 10.150          & 14        & 5.641  & 3.041 & 2.600  & 44.42 & 549.23 \\
40-75  & 40 & 40   & 2                & 9.400  & 76        & 9.550  & 0.000 & 9.550  & -1.60 & 88.59  \\
40-75  & 40 & 40   & 3                & 9.400           & 37        & 9.124  & 3.274 & 5.850  & 2.94  & 89.86  \\
40-75  & 40 & 40   & 4                & 9.400           & 25        & 8.549  & 3.749 & 4.800  & 9.05  & 92.12  \\
40-75  & 40 & 40   & 5                & 9.400           & 21        & 7.870  & 3.370 & 4.500  & 16.28 & 93.29  \\
40-75  & 40 & 50   & 2                & 9.400           & 59        & 8.947  & 1.547 & 7.400  & 4.82  & 114.26 \\
40-75  & 40 & 50   & 3                & 9.400           & 22        & 8.181  & 3.781 & 4.400  & 12.97 & 117.17 \\
40-75  & 40 & 50   & 4                & 9.400           & 14        & 7.140  & 3.690 & 3.450  & 24.04 & 120.61 \\
40-75  & 40 & 50   & 5                & 9.400           & 13        & 6.766  & 3.316 & 3.450  & 28.02 & 121.60 \\
40-75  & 40 & 60   & 2                & 9.400           & 27        & 8.425  & 3.775 & 4.650  & 10.37 & 143.99 \\
40-75  & 40 & 60   & 3                & 9.400           & 24        & 7.501  & 3.551 & 3.950  & 20.20 & 148.52 \\
40-75  & 40 & 60   & 4                & 9.400           & 16        & 6.714  & 3.164 & 3.550  & 28.57 & 152.56 \\
40-75  & 40 & 60   & 5                & 9.400           & 12        & 6.219  & 2.919 & 3.300  & 33.84 & 153.65 \\
40-75  & 40 & 70   & 2                & 9.400           & 24        & 7.937  & 3.937 & 4.000  & 15.56 & 178.91 \\
40-75  & 40 & 70   & 3                & 9.400           & 18        & 6.900  & 3.550 & 3.350  & 26.60 & 186.67 \\
40-75  & 40 & 70   & 4                & 9.400           & 14        & 6.018  & 2.918 & 3.100  & 35.98 & 191.19 \\
40-75  & 40 & 70   & 5                & 9.400           & 12        & 5.509  & 2.559 & 2.950  & 41.39 & 192.24 \\
40-75  & 40 & 80   & 2                & 9.400           & 42        & 7.450  & 2.500 & 4.950  & 20.74 & 206.34 \\
40-75  & 40 & 80   & 3                & 9.400           & 19        & 6.379  & 3.029 & 3.350  & 32.14 & 217.13 \\
40-75  & 40 & 80   & 4                & 9.400           & 12        & 5.700  & 3.300 & 2.400  & 39.36 & 222.69 \\
40-75  & 40 & 80   & 5                & 9.400           & 10        & 4.946  & 2.296 & 2.650  & 47.38 & 223.73
 \\ \hline
\end{tabular}}
\caption{Comparison between 2ER-TD and TSP modes}
\end{table}

\end{document}